\documentclass[11pt,a4paper]{article}
\usepackage[hyperref]{acl2018}
\usepackage{times}
\usepackage{latexsym}
\usepackage{amsmath}
\usepackage{algorithm}
\usepackage{algorithmic}
\usepackage{amsmath,bm}
\usepackage{amsfonts}
\usepackage{mathrsfs}
\usepackage{amsbsy,amsthm,amssymb}
\usepackage{graphicx}
\usepackage{xcolor}
\usepackage{multirow}
\usepackage{paralist}
\usepackage{subfigure}
\usepackage{arydshln}
\usepackage{CJK}
\usepackage{fancyhdr}
\usepackage{varwidth}
\usepackage{bbm}
\usepackage{url}
\usepackage{amsfonts}
\usepackage{amsmath}
\usepackage{multirow}
\usepackage{bm}
\aclfinalcopy 
\def\aclpaperid{} 

\begin{document}

\title{BFGAN: Backward and Forward Generative Adversarial Networks for \\ Lexically Constrained Sentence Generation}

\author{Dayiheng Liu$^\dagger$, Jie Fu$^\ddagger$, Qian Qu$^\dagger$, Jiancheng Lv$^\dagger$\thanks{\, Correspondence to Jiancheng Lv.}\\
$^\dagger$Data Intelligence Laboratory, Sichuan University \\
$^\ddagger$MILA, IVADO, Polytechnique Montreal\\
  {\tt losinuris@gmail.com
 }
 \\
  {\tt lvjiancheng@scu.edu.cn}
  } 
\maketitle

\date{}

\begin{abstract}

Incorporating prior knowledge like \textit{lexical constraints} into the model's output to generate meaningful and coherent sentences has many applications in dialogue system, machine translation, image captioning, etc. However, existing RNN-based models incrementally generate sentences from left to right via beam search, which makes it difficult to directly introduce lexical constraints into the generated sentences. In this paper, we propose a new algorithmic framework, dubbed BFGAN, to address this challenge. Specifically, we employ a backward generator and a forward generator to generate lexically constrained sentences together, and use a discriminator to guide the joint training of two generators by assigning them reward signals. Due to the difficulty of BFGAN training, we propose several training techniques to make the training process more stable and efficient. Our extensive experiments on two large-scale datasets with human evaluation demonstrate that BFGAN has significant improvements over previous methods.
\end{abstract}

\section{Introduction} \label{sec:intro}
In many real-world natural language generation scenarios, it is required to introduce \textit{lexical constraints} into the generated sequences, which is called \textit{lexically constrained} sentence generation \cite{Hokamp2017Lexically}. It generates a sentence that must contain specific words or phrases and is an active research topic in natural language generation. For examples, to avoid universal responses in the dialogue, a contextual keyword can be introduced into the reply \cite{Mou2016Sequence}. For machine translation, some specific domain terminologies may need to be introduced into the translation results \cite{Post2018Fast}. For image captioning, to mitigate the out-of-domain issue, image tag words can be incorporated into the output sentences \cite{Anderson2017Guided}.

In recent years, the recurrent neural network (RNN) based models have made remarkable progress in several natural language generation tasks, including neural machine translation \cite{wu2016google}, product review generation \cite{Dong2017Learning}, abstractive summarization \cite{See2017Get}, and affective text generation \cite{Ghosh2017Affect}, etc. However, existing RNN based models usually generate sentences incrementally from left to right by beam search \cite{och2004alignment}. It is difficult for these models to directly generate sentences containing lexical constraints. Replacing an arbitrary word in output with the desired word will damage the fluency of the sentence. Given words as additional inputs, there is no guarantee that they will appear in the outputs \cite{Anderson2017Guided,Yin2015Neural}. 

Various lexically constrained decoding methods \cite{Anderson2017Guided,Hokamp2017Lexically,Post2018Fast} have been proposed for lexically constrained sentence generation, some of which extends beam search to allow the inclusion of lexical constraints. However, these methods have high computational complexity \cite{Post2018Fast}. Moreover, these methods do not consider what specific words need to be included at the beginning of generation, but try to force specific words into sentences at every time step of the generation process. This unnatural way of generating may affect the quality of the generated sentence. We find that this problem is more serious when applying these methods for the unconditional language generative models.

Another more natural way of generating lexically constrained sentences is based on a Backward and Forward Language Model (B/F-LM) introduced in \cite{Mou2016Backward}, where a backward and a forward language model work together to generate lexically constrained sentences. The backward language model takes the lexical constraint as input to generate the first half of the sentence. Then the forward language model takes this first half as input to generate the whole sentence. These two language models are trained using maximum likelihood estimation (MLE) objectives. As argued in \cite{Bengio2015Scheduled}, they suffer from the exposure bias due to the discrepancy between training and inference stage: each language model generates a sequence conditioned on its previously generated ones during inference but they are trained to generate a sequence given ground-truth words.

More importantly, during preliminary experiments, we observed that the first half of the sentence generated by the backward language model and the second half of the sentence generated by the forward model tend to be semantically inconsistent or incoherent. This issue stems from the fact that the two language models are trained \textit{separately}, but are required to output together. In other words, there is no interaction between the two language models during training, and neither of them can get access to the outputs from each other. We believe a better way is to jointly train the two language models and let them get access to each other's output during training. 

To this end, we propose a new method, dubbed Backward-Forward Generative Adversarial Network (BFGAN), to address the above issues for lexically constrained sentence generation. Our method contains three modules: a \textbf{Backward Generator}, a \textbf{Forward Generator}, and a \textbf{Discriminator}. 
We extend the B/F-LM \cite{Mou2016Backward} in the following ways: In order to improve the coherence of the generation, we introduce the dynamic attention mechanism which is similar to the self-attention in \cite{Mei2016Coherent,vaswani2017attention} into the forward generator. With the dynamic attention mechanism, the scope of attention increases with the recursive operation from the beginning to the end of the sentence, and the forward generator can attend the first half sentence which is generated by the backward generator during inference. To solve the issue of discrepancies between training and inference in B/F-LM, we employ a discriminator to learn to distinguish real sentences and generated constrained sentences. This discriminator outputs reward signals to guide the joint training of the two generators. The two generators try to cooperate to generate lexically constrained sentences that can fool the discriminator. 

Following SeqGAN \cite{Yu2016SeqGAN} and LeakGAN \cite{Guo2017Long}, we model the lexically constrained sentence generation as a sequential decision making process and train the two generators with policy gradient. To our best knowledge, this is the first time that GANs have been used for generating lexically constrained sentences. Practically, it has been difficult to train GANs for text generation, especially to train two generator jointly for lexically constrained sentence generation. We propose several training techniques to make the training process more stable and efficient. With these training techniques, the BFGAN can be trained on large-scale datasets and generate long lexically constrained sentences (more than 20 words), and the BFGAN can be trained without additional labels.

We conduct extensive experiments on two large-scale datasets. For evaluation, we use BLEU score and self-BLEU score \cite{Zhu2018Texygen} to test the fluency and diversity of the generated sentences. In addition, we build a user-friendly web-based environment and launch a crowd-sourcing online study for human evaluation. In all cases, BFGAN shows significant improvements compared to the previous methods.

\section{Related Work}
\subsection{Lexically Constrained Decoding}
To incorporate pre-specified words or phrases in the output, one solution is lexically constrained decoding, which relies on a variant of beam search. For example, \cite{Hokamp2017Lexically} propose grid beam search to force certain words to appear in the output for machine translation. This method can be seen as adding an additional constrained dimension to the beam. The hypothesis must meet all constraints before they can be considered to be completed. To make existing image captioning models generalize to out-of-domain images containing novel scenes or objects, \cite{Anderson2017Guided} propose constrained beam search based on the finite-state machine, which is an approximate search algorithm capable of enforcing image tag over resulting output sequences. \cite{Post2018Fast} point out that these two methods are time-consuming, and their computation graphs are difficult to be optimized when running on GPUs. They propose a fast lexically-constrained decoding based on dynamic beam allocation (DBA), which can be seen as a fast version of grid beam search. 

Nevertheless, these methods still generate sentences from left to right. They do not consider what specific words need to be included at the beginning of generation, but try to force specific words into sentences at every time step of the generation process, which may affect the quality of the generated sentence. We find that this problem is more serious when applying these methods for the unconditional language generative models. Because our method starts with the given word, it has already known the lexical constraint before the generation, which is a more natural way. Our experimental results show that the BFGAN significantly outperforms the grid beam search \cite{Hokamp2017Lexically} based unconditional language generative models.

\subsection{Backward and Forward Language Modeling} \label{sec:bf}
To generate lexically constrained sentences, \cite{Mou2016Backward} first propose three variant backward and forward language models: separated, synchronized, and asynchronously backward and forward language model (called sep-BF, syn-BF, and asyn-BF). Their experiments show that asyn-BF is the most natural way of modeling forward and backward sequences. \cite{Mou2016Sequence} incorporate the asyn-BF model into the seq2seq framework to generate conversation reply containing the given keyword. 

The way BFGAN generates lexically constrained sentences is similar to asyn-BF as discussed in Introduction section. However, as discussed above, training backward and forward models separately with MLE will suffer from serious exposure bias, especially for unconditional language models. The two models in asyn-BF have never been trained together. The essential difference of our method is that it incorporates a discriminator to guide the joint training of the two generators. The experimental results show that BFGAN performs constantly better than state-of-the-art models.

\subsection{GANs for Text Generation}
Applying adversarial learning \cite{Goodfellow2014Generative} to text generation has recently drawn significant attention, which is a promising framework for alleviating the exposure bias issue \cite{huszar2015not}. \cite{Yu2016SeqGAN} take the sequence generation as a sequential decision making process. They propose SeqGAN which uses the output of the discriminator as reward and trains generator with policy gradient. \cite{Li2017Adversarial} extend SeqGAN model for dialogue generation. Instead of training binary classification as discriminator, \cite{Lin2017Adversarial} propose RankGAN that uses a ranker to make better assessment of the quality of the samples. MaliGAN \cite{Che2017Maximum} uses importance sampling combined with the discriminator output to deal with the reward vanishing problem. To address the sparse reward issue in long text generation, \cite{Guo2017Long} propose LeakGAN which leaks the feature extracted by the discriminator to guide the generator.

BFGAN is distinct in that it employs two generators for lexically constrained sentence generation, and employs a discriminator to guide the joint training of the two generators. In order to solve the problem of difficult training of the model. We propose several training techniques to make the training process more stable and efficient. To the best of our knowledge, this is the first time a GAN has been successfully used for generating lexically constrained sentences.

\section{Methodology}

\subsection{Overall Architecture} \label{section:overall}
\begin{figure}[t] 
   \centering
   \includegraphics[width=2.0in]{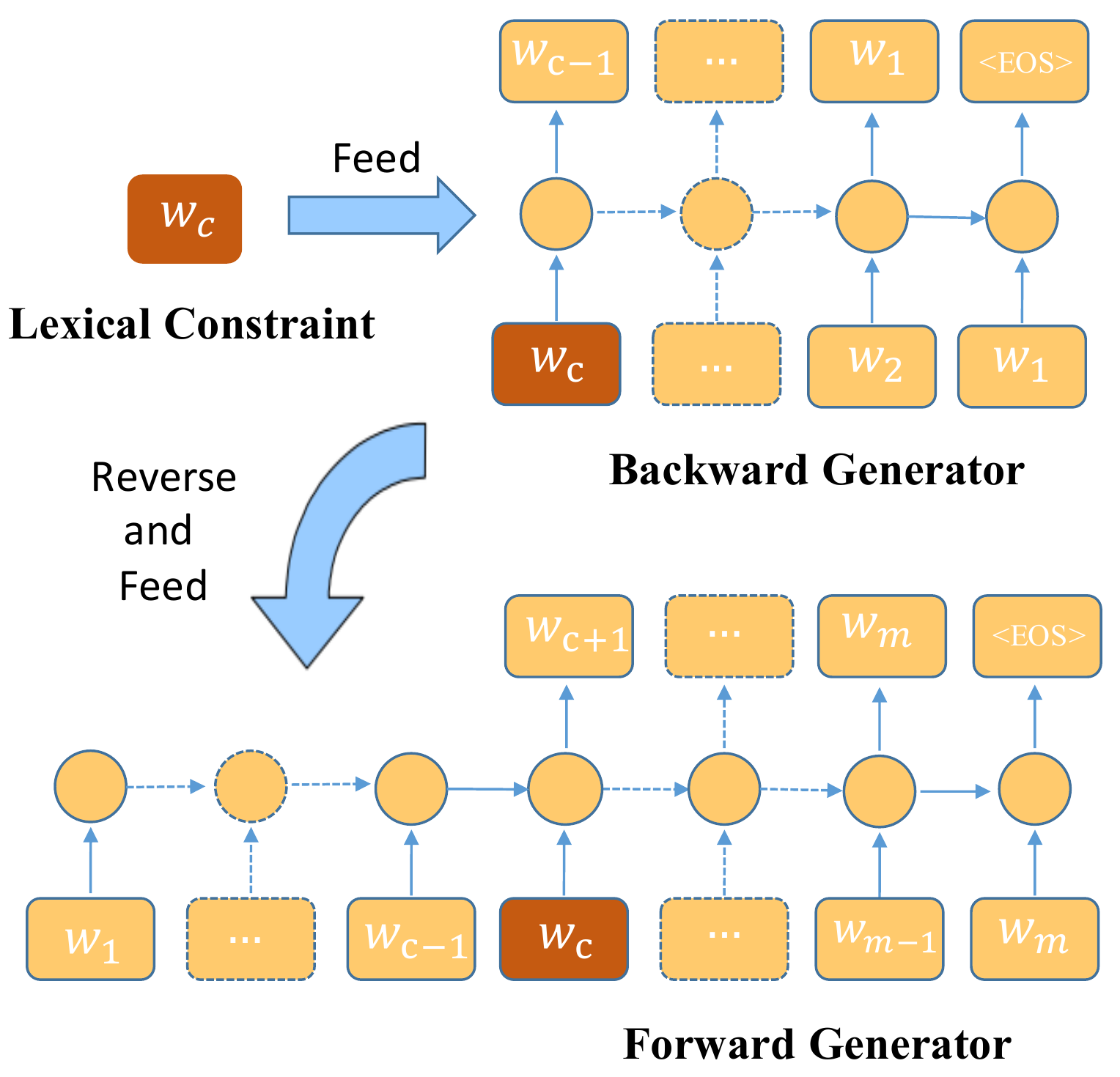}
   \caption{The generation process of backward and forward generators.}
   \label{fig:overall}
\end{figure}

The process of generation is shown in Figure \ref{fig:overall}, which involves three modules:

\noindent \textbf{Discriminator}. It learns to distinguish real sentences from machine-generated lexically constrained sentences. It guides the joint training of two generators by assigning them reward signals.

\noindent \textbf{Backward Generator}. Given a lexical constraint\footnote{It can be a word, a phrase or a sentence fragment.}, the backward generator takes it as the sentence's starting point, and generates the first half sentence backwards. 

\noindent \textbf{Forward Generator}. The sequence produced by the backward generator is reversed and fed into the forward generator. It then learns to generate the whole sentence with the aim of fooling the discriminator.

\subsection{Backward and Forward Generators} \label{section:generator}
Let $w_c$ denote the given lexical constraint. We denote the generated lexically constrained sentence with length $m$ as $s=w_1\cdots w_c\cdots w_m$\footnote{Note that $w_c$ can also appear at the beginning, or the end.}. The backward half sentence $s_{<c} = w_{c-1} \cdots w_1$ is generated by the backward generator $G^{(\rm bw)}_\theta$, and the forward half sentence $s_{>c} = w_{c+1} \cdots w_m$ is generated by the forward generator $G^{(\rm fw)}_{\theta'}$. $\theta$ and $\theta'$ represent the parameters of the backward generator and the forward generator respectively. The joint probability of the sentence can be written as:
\begin{flalign}
G(s|w_c;\theta,\theta') = P^{\text{(bw)}}_\theta(s_{<c}|w_c) P^{\text{(fw)}}_{\theta'}(s_{>c}|s_{1:c}),  
\end{flalign}
where $s_{1:c} = w_1 \cdots w_c $. 

The backward generator $G^{(\rm bw)}_\theta$ models the probability of the backward half sentence:
\begin{align}
P^{\text{(bw)}}_\theta(s_{<c}|w_c) =  \prod_{i=1}^{c-1} P^{\text{(bw)}}_\theta(w_{c-i}|w_c \cdots w_{c-i+1}), 
\end{align}

The forward generator $G^{(\rm fw)}_{\theta'}$ models the probability of the forward half sentence:
\begin{align}
P^{\text{(fw)}}_{\theta'}(s_{>c}|s_{1:c}) = \prod_{i=1}^{m-c} P^{\text{(fw)}}_{\theta'}(w_{c+i}|w_1 \cdots w_{c+i-1}). 
\end{align}
The two generators have the same structure but have distinct parameters. To improve the coherence of the constrained sentence, we employ an LSTM-based language model with dynamic attention mechanism (called \textbf{attRNN-LM}) as generator. This attention mechanism is similar to attention in RNN-Seq2Seq \cite{Bahdanau2014Neural}, but its scope of attention increases dynamically with the recursive operation from the beginning to the end of the sentence. In contrast, the RNN-Seq2Seq's attention scope is fixed to the hidden states of the encoder.

\begin{figure}[t] 
  \centering
  \subfigure[attoRNN-LM.]
  {
  \begin{minipage}{3.7cm} \label{fig:atto}
  \centering
    \includegraphics[width=3.5cm]{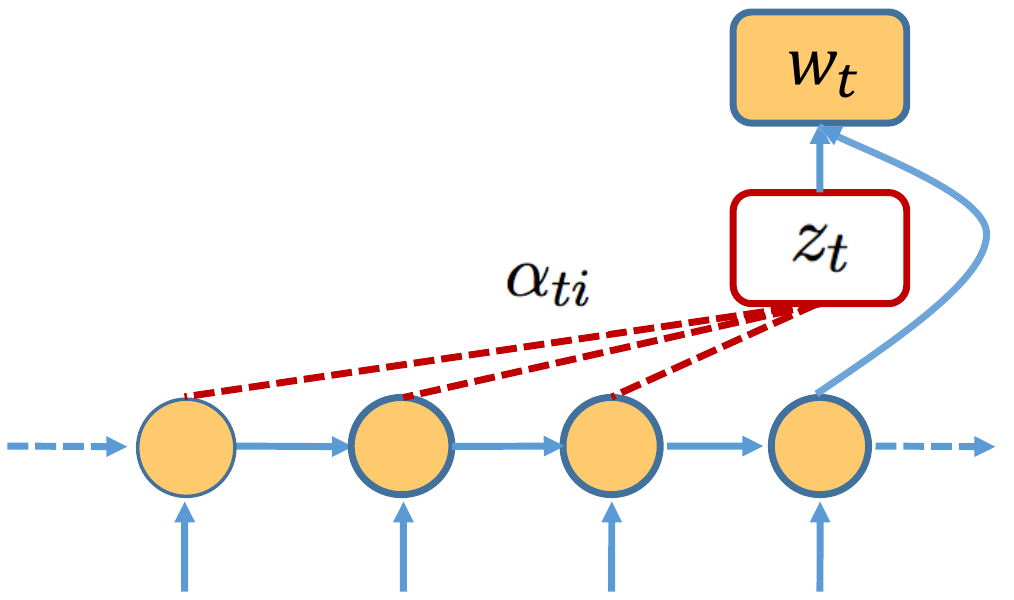}
  \end{minipage}
  }
  \subfigure[attinRNN-LM.]
  {
  \begin{minipage}{3.7cm}\label{fig:attin}
  \centering
    \includegraphics[width=3.5cm]{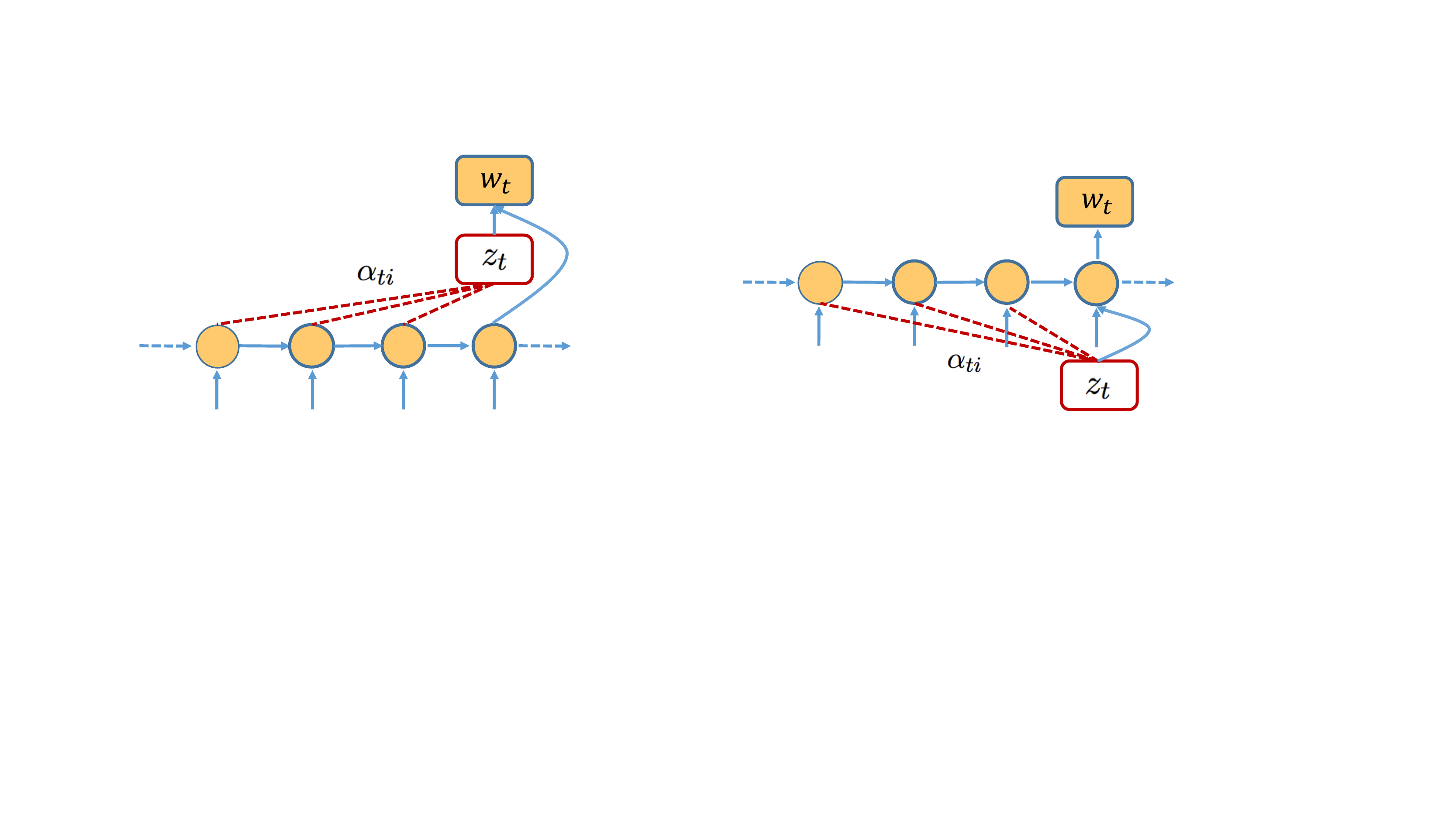}
  \end{minipage}
  }
  \caption{The architecture of two variants of attRNN-LM. }
 \label{fig:att}
 \end{figure}

The architecture of attRNN-LM is shown in Figure \ref{fig:att}. As with the RNN-LM, the model encodes the input sequence $w_1 \cdots w_t$ into a sequence of hidden states $ h_1 \cdots h_t$. The attention context vector $z_t$ at step $t$ is computed as a weighted average of history hidden states $h_{0:t-1}$: 
\begin{align}
        & e_{ti} = v^{\rm T} {\rm tanh}(W^a h_{t-1}+U^a h_i), \\
	& \alpha_{ti} = \frac{{\rm exp}(e_{ti})}{\sum_{k=0}^{t-1}{{\rm exp}(e_{tk})}}, \\
	& z_t= \sum_{i=0}^{t-1}{\alpha_{ti}{h}_i}. 
\end{align} 
There are two variants of attRNN-LM: the first one called \textbf{attoRNN-LM} (see Figure \ref{fig:atto}) is similar to A-RNN in \cite{Mei2016Coherent}, which uses the context vector $z_t$ together with hidden state $h_t$ to predict the output at time $t$:
\begin{align}
 h_t & = {\rm LSTM}\left (w_{t-1}, h_{t-1}\right ), \\ 
 P(w_t | s_{<t}) & = {\rm Softmax}\left (O^{\rm T} (W^h h_t + W^z z_t)) \right ). 
\end{align}

Another variant called \textbf{attinRNN-LM} (see Figure \ref{fig:attin}) uses the linear combination of $z_t$ and $w_{t-1}$ as input to RNN to get the hidden state $s_t$ and output at time $t$:
\begin{align}
g_t  & = W^e w_{t-1} + W^z z_t \\
 h_t & = {\rm LSTM}\left(g_{t}, h_{t-1}\right), \\
 P(w_t | s_{<t}) & = {\rm Softmax}\left(O^{\rm T} h_t \right). 
\end{align}
The attRNN-LM allows for a flexible combination of history generated words for every individual word. The forward generator can attend the first half sentence which is generated by the backward generator during inference. 
Empirically, we find that both the attinRNN-LM and attoRNN-LM perform better than RNN-LM for our problems.

\subsection{Discriminator} \label{section:discriminator}
The discriminator $D_\phi$ learns to distinguish real sentences and generated lexically constrained sentences. It guides the joint training of two generators by assigning proper reward signals. This module can be a binary classifier or a ranker \cite{Lin2017Adversarial}. We use CNN based model \cite{kim2014convolutional} as the discriminator which outputs a probability indicating whether the input is generated by humans or machines in the experiments.\footnote{We also tried LSTM-based discriminators but found CNN-based model is more effective and efficient.}

\subsection{Training} \label{section:train}
\subsubsection{Pre-training}
The two generators are firstly pre-trained by standard MLE loss. Note that our method can be trained without additional labels. Given a sentence, we randomly sample a notional word for it as the lexical constraint, then we take the first half and reverse its word order as a training sample for backward generator. We repeat this way to construct multiple training samples for each real sentence. The forward model is trained the same as the standard RNN language model. 

After pre-training the generators, we pre-train the discriminator by providing positive examples from the human-written sentences and negative examples generated from the generators. Given a positive sample $s$, we construct multiple negative samples $\hat{s}$ in the following ways: 1) We slice a random sentence fragment in the middle of the sample $s$, and take this sentence fragment as a negative sample. For example, given a sentence ``my pastor recommended this book, and it was really insightful.'', we may get a sentence fragment like this: ``recommended this book, and it was really''; 2) We take the sentence fragment as the lexical constraint for BFGAN to generate a complete sentence by beam search; 3) We use BFGAN to generate a complete sentence containing the lexical constraint from sampling; 4) We use the backward generator to generate a half sentence without any constraint by sampling and the forward generator to generate a complete sentence by sampling. By constructing different types of strong negative samples and weak negative samples, the generalization ability of the discriminator is increased. Therefore it can provide better reward signals for the generators. We train the discriminator using gradient ascend of the discriminator loss:
\begin{align}
J(\phi) = \log D_\phi(s) + \log (1-D_\phi(\hat{s})). 
\end{align}

\subsubsection{Policy Gradient Training}
The discriminator's output $D_\phi(\hat{s})$ is the probability that the generated sentence $\hat{s}$ is written by human. We use $D_\phi(\hat{s})$ as the reward $r$ to encourage the two generators to work together to generate sentences which are indistinguishable from human-written sentences. 
The generators are trained by the REINFORCE algorithm \cite{Williams1992Simple}: 
\begin{align} \label{eq:pg}
J(\theta, \theta') = \mathbb{E}_{s\sim G(s|w_c;\theta,\theta')} D_\phi(s). 
\end{align}
The gradient of Eq. (\ref{eq:pg}) is approximated using the likelihood ratio trick \cite{Sutton1999Policy,Glynn1990Likelihood}:
\begin{align}
\nabla J(\theta, \theta')  \approx (D_\phi(s) - b) \nabla {\rm log}G(s| w_c; \theta,\theta'),  
\end{align}
where $b$ is the baseline value to reduce the variance of the estimate.

\subsubsection{Training Techniques}
To construct the lexical constraints during policy gradient training, we randomly sample a batch of real sentence at each step. For each real sentence $s$ with length $m$, we slice it from the middle into a sentence fragment $s_{c}$ at length $m-\Delta \times i$. This sentence fragment is taken as the lexical constraint and fed into the generators to generate a fake example. 

In order to gradually teach the model to produce stable sentences, we provide more and more difficult lexical constraints, which is a form of curriculum learning \cite{bengio2009curriculum}. We increase the $i$ to make the lexical constraints shorter and shorter. For generators, the part of lexical constraint $s_{c}$ is trained with cross entropy loss and the remaining generated words are trained with policy gradient. To make the training stable and prevent the perplexity value skyrocketing, we apply teacher forcing to giving the generators access to the gold-standard targets after each policy training step.

Monte Carlo (MC) search is used to estimate the reward for every generation step in \cite{Yu2016SeqGAN}. However this sampling process is time-consuming especially for long sentence generation \cite{Li2017Adversarial}, thus making it hard to be applied to large-scaled datasets. In our experiment, given the lexical constraint, the generators use beam search to generate the whole sentence $\hat{s}$, which is consistent with the inference stage. Then we get the reward $D_\phi(\hat{s})$ for each generated word from the discriminator. We empirically found this technique can improve the efficiency and stability of training.

The details of training procedure are summarized in Algorithm \ref{algo}.

\vspace{0.2cm}
\begin{algorithm}
\caption{Training Algorithm of BFGAN}
{\fontsize{9}{9}\selectfont
\begin{algorithmic}[1]
\REQUIRE  \begin{varwidth}[t]{\linewidth}
A backward generator $G^{(\rm bw)}_{\theta}$ with parameters $\theta$. \par
A forward generator $G^{(\rm fw)}_{\theta'}$ with parameters $\theta'$. \par
A discriminator $D_\phi$ with parameters $\phi$.\par
A baseline $b$, step size $\Delta$, maximum length $T$. \par
\end{varwidth} 
\vspace{2mm}
\STATE Pre-train the $G^{(\rm fw)}_{\theta'}$, $G^{(\rm bw)}_{\theta}$ and $D_\phi$ for certain epochs.
\FOR{$K$ = $T$, $1$, $-\Delta$}
\FOR{number of training iterations}
    \STATE Sample $s$ from the dataset. Get the lexical constraints $s_c$ with length $K$.
    \STATE Given $s_c$, let $G^{(\rm fw)}_{\theta'}$ and $G^{(\rm bw)}_{\theta}$ generate the whole sentence $\hat{s}$ using beam search.
    \STATE Compute the reward $r$ of $\hat{s}$ using $D_\phi$.
    \STATE Update $G^{(\rm bw)}_{\theta}$ on $\hat{s}_{<c}$ with reward $r$.
    \STATE Update $G^{(\rm fw)}_{\theta'}$ on $\hat{s}_{>c}$ with reward $r$.
    \STATE Update $G^{(\rm bw)}_{\theta}$ and $G^{(\rm fw)}_{\theta'}$ on $s_c$ using cross entropy.
    \STATE Teacher-Forcing: Update $G^{(\rm bw)}_{\theta}$ and $G^{(\rm fw)}_{\theta'}$ on $s$.
    \FOR{$i=1$, D-steps}
    \STATE Sample $s$ from real data, construct the negative samples $\hat{s}$.
    \STATE Update $D_\phi$ using $s$ as as positive samples and $\hat{s}$ as negative samples.
    \ENDFOR
    \ENDFOR
\ENDFOR
\end{algorithmic}
}
\label{algo}
\end{algorithm}

\section{Experiments}
The experiments are designed for answering the following questions: \textbf{Q1:} Since we employ two variants attRNN-LM as the generators, compared with RNN-LM, which model is the most effective one? \textbf{Q2:} What is the performance of the model after training with the proposed adversarial learning. \textbf{Q3:} Judging from the human views, how does the proposed approach compare with the existing methods?

\subsection{Datasets}
Two large-scale datasets are used here. The first one is the Amazon Product Reviews Corpus (APRC) \cite{Dong2017Learning} which is built upon Amazon product data \cite{mcauley2015inferring}. We set the vocabulary size to 10K, and the maximum length of the sentences to 35. After preprocessing, the first corpus contains 350K review sentences.

The second dataset is Chinese Sentence Making Corpus (CSMC).\footnote{In order to make our results easy to reproduce, we will release all the datasets and codes upon acceptance of the paper.} 
We manually crawl and build this dataset which contains 107K Chinese common words or idioms. Each common word or idiom have several example sentences which contain that word or idiom. There are 1.34M example sentences. The maximum length of these sentences is 30. Compared with the APRC dataset, the CSMC is a more open-domain dataset and training the model on it will be more challenging. These two datasets are randomly split into train/valid/test sets following these ratios respectively: 85\%, 5\%, 10\%.

\subsection{Structure Comparison of Generator (Q1)}
To answer the first question, we train the forward generators with RNN-LM, attinRNN-LM, and attoRNN-LM structures on CSMC dataset separately. All the generators are 2-layer char-level LSTMs with 1024 hidden units. The dimension of word embeddings is set to 1024. The batch size, dropout rate, threshold of element-wise gradient clipping and initial learning rate of Adam optimizer \cite{kingma2014adam} are set to 128, 0.5, 5.0 and 0.001. Layer Normalization \cite{ba2016layer} is also applied. 

We use the perplexity (PPL) metric to evaluate the generators. In addition, we use BLEU score as an additional evaluation metric. Following \cite{Yu2016SeqGAN,Lin2017Adversarial}, we randomly sample 10,000 sentences from the test set as the references. For each generator, we let it generate 10,000 sentences by sampling. We report the average BLEU-4 scores of the generated sentences. The results are shown in Table \ref{exp:1}. From the results we can see that both attinRNN-LM and attoRNN-LM perform better than RNN-LM, and attinRNN-LM achieves the best performance. In the following experiments, we use the attinRNN-LM structure as the backward and the forward generators.

\begin{table}[h]
\small
\begin{center}
\begin{tabular}{lcccc}
 \hline
 Models &  PPL (dev) & PPL (test) & BLEU-4\\
 \hline \hline
RNN-LM & 17.95 & 18.15 & 0.193 \\ 
attoRNN-LM  & 16.18 & 16.37 & 0.219\\ 
attinRNN-LM  & \textbf{15.81} & \textbf{16.07} & \textbf{0.225}\\  
 \hline
 \end{tabular}
\end{center}
\caption{\label{exp:1} The evaluation results of different generators on CSMC.}
\end{table}

\begin{table}[h]
\small
\begin{center}
\begin{tabular}{lcccc}
 \hline
 Models &  BLEU-2 & BLEU-3 & BLEU-4\\
 \hline \hline
B/F-LM & 0.799 & 0.476 & 0.190 \\
BF-MLE & 0.798 & 0.482 & 0.203 \\ 
BFGAN  & 0.836 & 0.543 & 0.248\\  
BFGAN-Mali  & 0.829 & 0.536 & 0.246\\ 
BFGAN-rescale  & \textbf{0.850} & \textbf{0.574} & \textbf{0.295}\\  
 \hline
  \hline
 Models &  sBLEU-2 & sBLEU-3 & sBLEU-4\\
 \hline \hline
B/F-LM & 0.807 & 0.498 & 0.236 \\ 
BF-MLE & 0.810 & 0.516 & 0.269 \\ 
BFGAN  & 0.852 & 0.624 & 0.399\\  
BFGAN-Mali  & 0.849 & 0.617 & 0.392\\ 
BFGAN-rescale  & 0.849 & 0.644 & 0.445\\  
 \hline
 \end{tabular}
\end{center}
\caption{\label{exp:bleu-csmc} BLEU and self-BLEU (sBLEU) scores on CSMC.}
\end{table}

\begin{table}[h]
\small
\begin{center}
\begin{tabular}{lcccc}
 \hline
 Models &  BLEU-2 & BLEU-3 & BLEU-4\\
 \hline \hline
B/F-LM & 0.843 & 0.653 & 0.527 \\
BF-MLE & 0.854 & 0.665 & 0.539 \\ 
BFGAN  & 0.920 & 0.796 & 0.672\\  
BFGAN-Mali  & 0.920 & 0.803 & 0.675\\ 
BFGAN-rescale  & \textbf{0.926} & \textbf{0.814} & \textbf{0.698}\\  
 \hline
  \hline
 Models &  sBLEU-2 & sBLEU-3 & sBLEU-4\\
 \hline \hline
B/F-LM & 0.924 & 0.811 & 0.680 \\
BF-MLE & 0.925 & 0.810 & 0.681 \\ 
BFGAN  & 0.937 & 0.837 & 0.721\\  
BFGAN-Mali  & 0.936 & 0.834 & 0.717\\ 
BFGAN-rescale  & 0.937 & 0.840 & 0.729 \\  
 \hline
 \end{tabular}
\end{center}
\caption{\label{exp:bleu-aprc} BLEU and self-BLEU (sBLEU) scores on APRC.}
\end{table}

\subsection{Evaluation of BFGAN (Q2)}
In this experiment, we train the BFGAN on both CSMC and APRC datasets. The hyperparameters of the backward and the forward generators trained on CSMC are the same as those in the first experiment. Since the APRC dataset is smaller than CSMC, we set both backward and forward generators to one layered word-level LSTM with 1024 hidden units when training on APRC.

We take the BF-MLE (the backward and the forward generators without adversarial learning), and the B/F-LM (the backward and forward LSTM based language models) \cite{Mou2016Backward} as the baselines. In addition, we compare the basic BFGAN with BFGAN-rescale (the BFGAN which use Bootstrapped Rescaled Activation \cite{Guo2017Long} to rescale the reward during training), and the BFGAN-Mali (the BFGAN that use the same reward rescaling as in the MaliGAN \cite{Che2017Maximum} during training). To be fair, all models use roughly the same number of parameters.

As with the evaluation in \cite{Zhu2018Texygen}, we use BLEU score to measure the similarity degree between the generated sentences and the real sentences. We randomly sample 10,000 sentences from the test set as the references. Moreover, we use self-BLEU score to evaluate the diversity of the generated sentences. For each model, we let its backward generator generate 10,000 half sentences and its forward generator generate complete sentences from sampling.

The evaluation results are shown in Table \ref{exp:bleu-csmc} and Table \ref{exp:bleu-aprc}. Firstly, we can see that the BF-MLE is better than B/F-MLE. This result shows the effectiveness of attinRNN-LM again. Secondly, we find that the BLEU-4 score of BF-MLE is lower than the forward generator (see Table \ref{exp:1}). In other words, the fluency of the sentences generated by the two generators together is worse than that generated by the forward generator only. These results prove what we mentioned above, the two generators trained \textit{separately} indeed suffer from more serious exposure bias than one generator. However, after training the two generators with the proposed adversarial learning, the BLEU scores of all BFGAN variants are much higher than those of BF-MLE and B/F-LM on both datasets. These results show that BFGAN performs much better than B/F-LM. In addition, the BFGAN-rescale achieves the highest BLEU scores.

From the results of self-BLEU score, compared with other models, we find that the BFGAN-rescale suffers from more serious mode collapse \cite{arjovsky2017wasserstein}. On both datasets, we can see that the self-BLEU scores of BF-MLE and B/F-LM are lower than those of other models. The BF-MLE and B/F-LM can generate sentences with higher diversity. These results are in line with the evaluation in \cite{Zhu2018Texygen}, the self-BLEU score of various text GAN is higher than that of model trained by MLE. We can see that the self-BLEU scores on APRC dataset of BFGAN are close to those of BF-MLE and B/F-LM, but for CSMC dataset, the self-BLEU scores of B/F-LM are much lower than those of BFGAN variants. We further analyzed this reason. Because the CSMC dataset is a more open-domain dataset, the B/F-LM surfers from more serious exposure bias on this dataset and often generates unreal samples. These low quality samples may have greater diversity. 

\begin{figure*}[h] 
   \centering
   \includegraphics[width=6.0in]{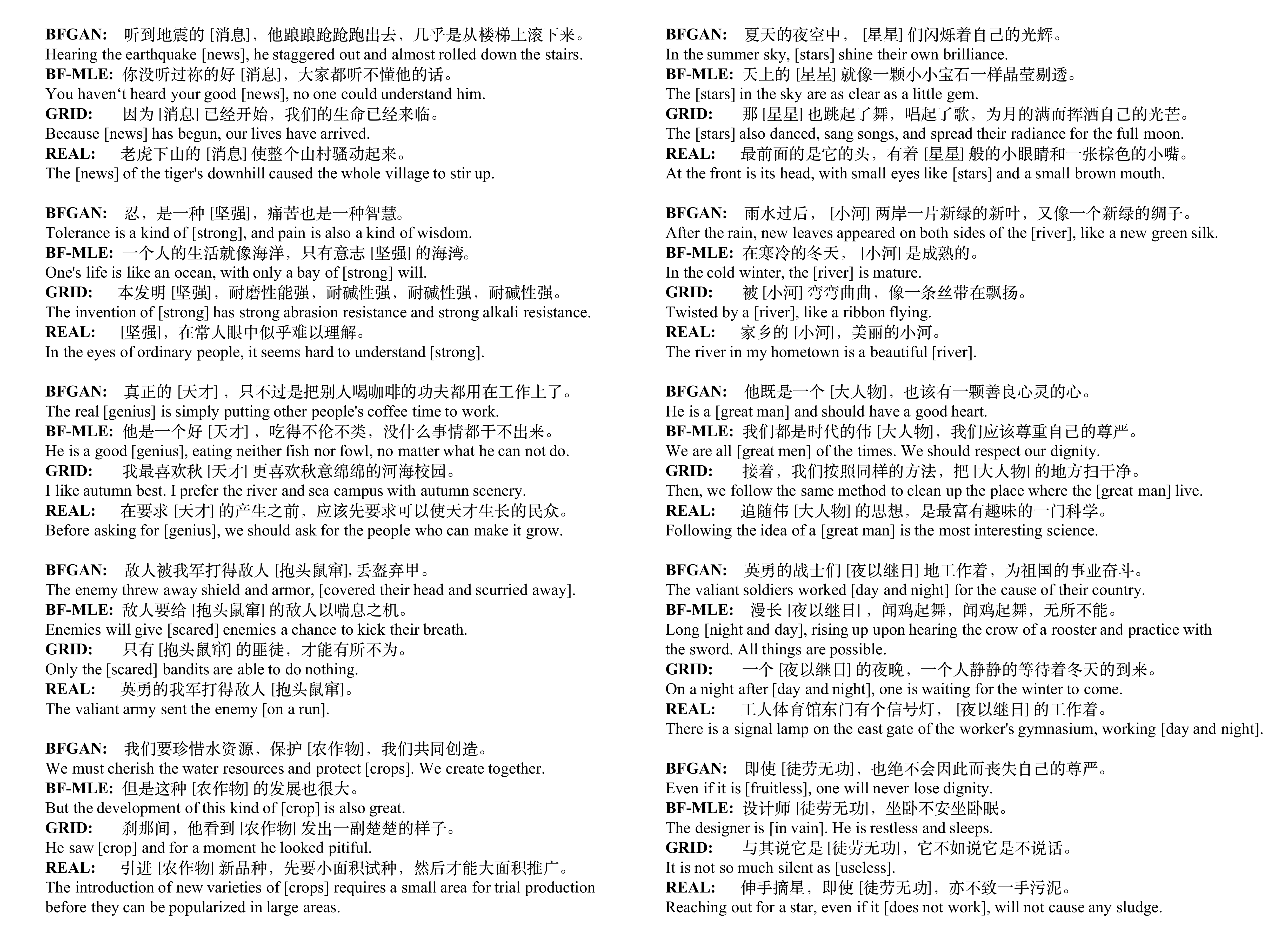}
   \caption{Some generated examples. GRID: grid beam search, REAL: real samples. We use [$\cdot$] to highlight the lexical constraints. Since the original generated sentences are Chinese, we manually translate them into English alongside. }
   \label{fig:example}
\end{figure*}

\subsection{Human Evaluation (Q3)}
To further evaluate the proposed method, we conduct several human evaluations. In this experiment, we compare the BFGAN with BF-MLE and grid beam search \cite{Hokamp2017Lexically}. In addition, we compare BFGAN with human, which can be regarded as a Turing test. All comparisons are blind paired comparisons.

We randomly pick 50 common words or idioms as lexical constraints, and let all models generate a lexically constrained sentence for each lexical constraint. For grid search, we use forward generator with grid beam search algorithm to force the lexical constraint appear in the output. To be fair, all models are the same structure. For each lexical constraint, we randomly pick a real sample that contains this constraint from the dataset as a human-written sample. Then we launch a crowd-sourcing online study asking evaluators to decide which generated sentence is better. We build a user-friendly web-based environment for this experiment, and the interface for human interaction is illustrated in Appendix. For each round, the human evaluation interface presents two sentences which are generated by two different methods with the same lexical constraint, then asks evaluators to choose the better one. Ties are permitted. A total of 50 evaluators\footnote{All evaluators are well educated and have Bachelor or higher degree. They are independent of the authors' research group.} participate in the evaluation. 

Table \ref{exp:human-csmc} presents the results on CSMC dataset. Only 4\% sentences generated by the grid beam search and 6\% sentences generated by BF-MLE were considered better than those produced by the BFGAN. We can observe a significant quality improvement from the BFGAN. From the result of comparison between BFGAN and human-written samples, we can see that 56\% sentence groups make the evaluators unable to tell which is better. Surprisingly, there is 6\% sentences generated by the BFGAN are considered even better than human-written samples. These results show that our method can generate high quality lexically constrained sentences.

\begin{table}[h]
\small
\begin{center}
\begin{tabular}{cccc}
 \hline
BFGAN won & BF-MLE won & Tied\\
 \hline 
\textbf{76\%} & 6\% &18\% \\  
\hline\hline
BFGAN won & Grid Beam Search won & Tied\\
 \hline \hline
\textbf{82\%} & 4\% &14\% \\ 
\hline\hline
BFGAN won & Human won & Tied\\
 \hline 
6\% & 38\% & \textbf{56\%} \\ 
\hline
 \end{tabular}
\end{center}
\caption{\label{exp:human-csmc} Human evaluations on CSMC. The comparison result between BFGAN and BF-MLE is shown at the top. The comparison result between BFGAN and grid beam search is shown in the middle. At the bottom of the table is the comparison result between BFGAN and human-written samples.}
\end{table}

\begin{figure}[t] 
  \centering
  \subfigure[]
  {
  \begin{minipage}{6.0cm} \label{fig:visual1}
  \centering
    \includegraphics[width=6.0cm]{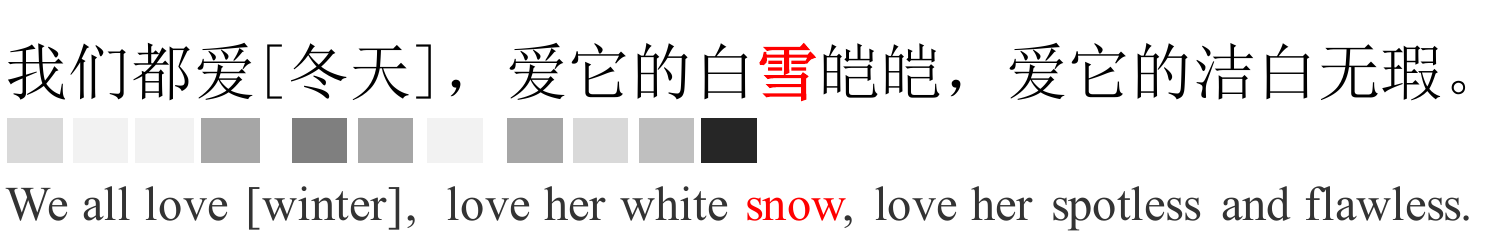}
  \end{minipage}
  }
  \subfigure[]
  {
  \begin{minipage}{6.0cm}\label{fig:visual2}
  \centering
    \includegraphics[width=6.0cm]{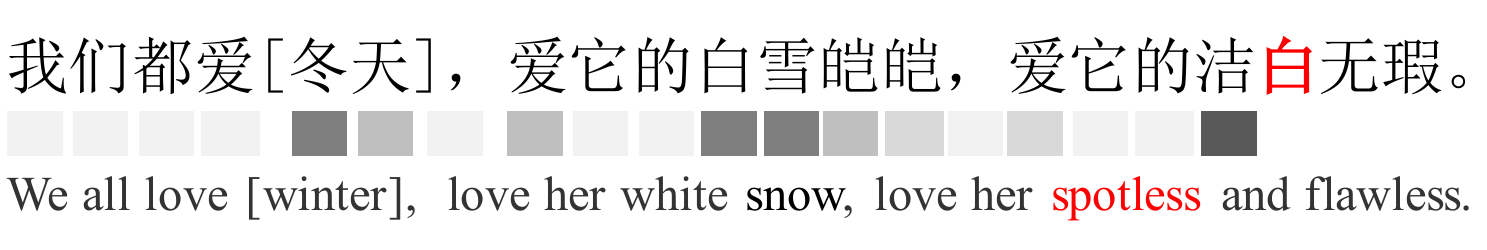}
  \end{minipage}
  }
  \caption{Visualization of attentions. The blocks under the words represent the attention weights.}
 \label{fig:visual}
 \end{figure}

\section{Qualitative Analysis}
We show some evaluation examples in Figure \ref{fig:example}. We can see that the BFGAN can generate high-quality lexical constrained sentences. Some of them are even indistinguishable from sentences written by human. While most sentences generated by BF-MLE are lack of coherence and fluency. In addition, the quality of the sentence generated by grid search is the worst. Since we introduce the dynamic attention mechanism into the generators, we qualitatively evaluate the effectiveness of it through the visualizations of the attention weights. Figure \ref{fig:visual} provides two visualization samples of the attention weights in the BFGAN. Figure \ref{fig:visual} (a) and (b) present the generated sentences and we highlight in red two output words for two time steps. For each highlighted generated word, we visualize the attention weights of each previous generated word, where darker shades indicate larger attention weights. We can see that the dynamic attention mechanism helps learn a better generator that promotes the coherence, by learning to associate the currently-generated word with informative contextual words. We show some examples of APRC dataset in Appendix.

\section{Conclusion and Future Work}

In summary, our key contributions are as follows: 
\begin{itemize}
    \item We propose a new method, called BFGAN, to address the challenge of generating sentences with lexical constraints. BFGAN significantly improves the quality of lexically constrained sentence generation compared with state-of-the-art methods.
    \item To ease the training of BFGAN, we propose an efficient and stable training algorithm. Furthermore, it does not need additional labels.
    \item Extensive experiments based on two large-scale datasets and human evaluations demonstrate significant improvements over previous methods.
\end{itemize}

For future work, we plan to extend our method to sequence-to-sequence frameworks and apply it to more nature language generation applications like dialogue systems. It might also be interesting to explore how to generate sequences with multiple lexical constraints. 

\section*{Acknowledgment}
This work was supported by the National Science Foundation of China (Grant No.61625204), partially supported by the State Key Program of National Science Foundation of China (Grant No.61432012 and 61432014).

\bibliographystyle{acl_natbib}
\bibliography{bfgan}

\newpage

\section{Appendix}
\subsection{Interface of Human Evaluation}
\begin{figure}[h] 
  \centering
   \includegraphics[width=2.7in]{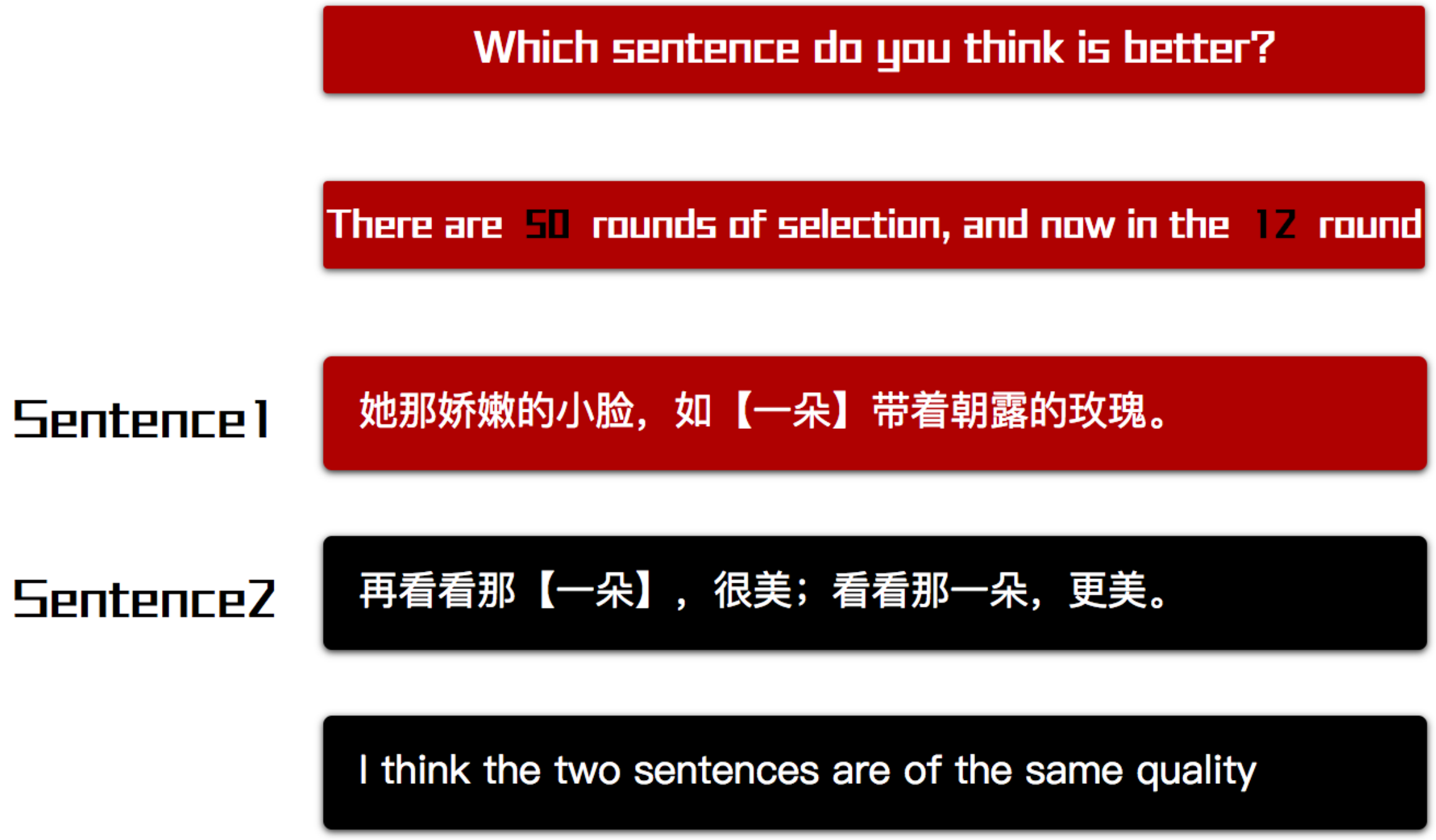}
   \caption{The human evaluation interface presents two sentences which are generated by two different methods with the same lexical constraint at a time. It asks users to choose the better one. Ties are permitted.}
   \label{fig:interface}
\end{figure}

We build a user-friendly web-based environment based on Flask\footnote{http://flask.pocoo.org/} for human evaluation experiment. The interface for human interaction is illustrated in Figure \ref{fig:interface}.

\begin{figure}[h] 
   \centering
   \includegraphics[width=3in]{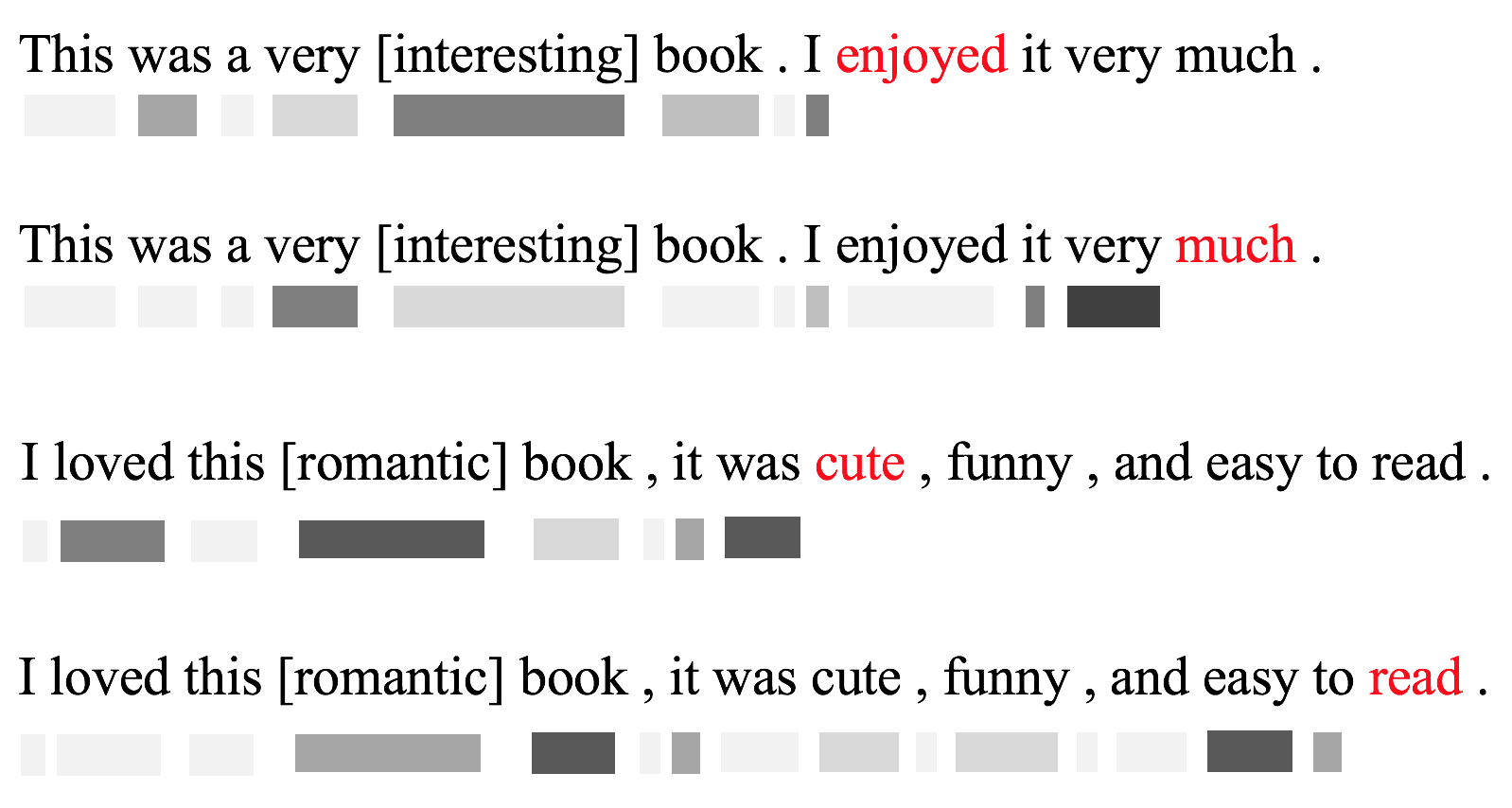}
 \caption{Visualization of attentions. The blocks under the words represent the attention weights. Darker shades indicate larger attention weights. We use [$\cdot$] to highlight the lexical constraints.}
 \label{fig:visual_aprc}
\end{figure}

\begin{figure}[h] 
   \centering
   \includegraphics[width=3.9in]{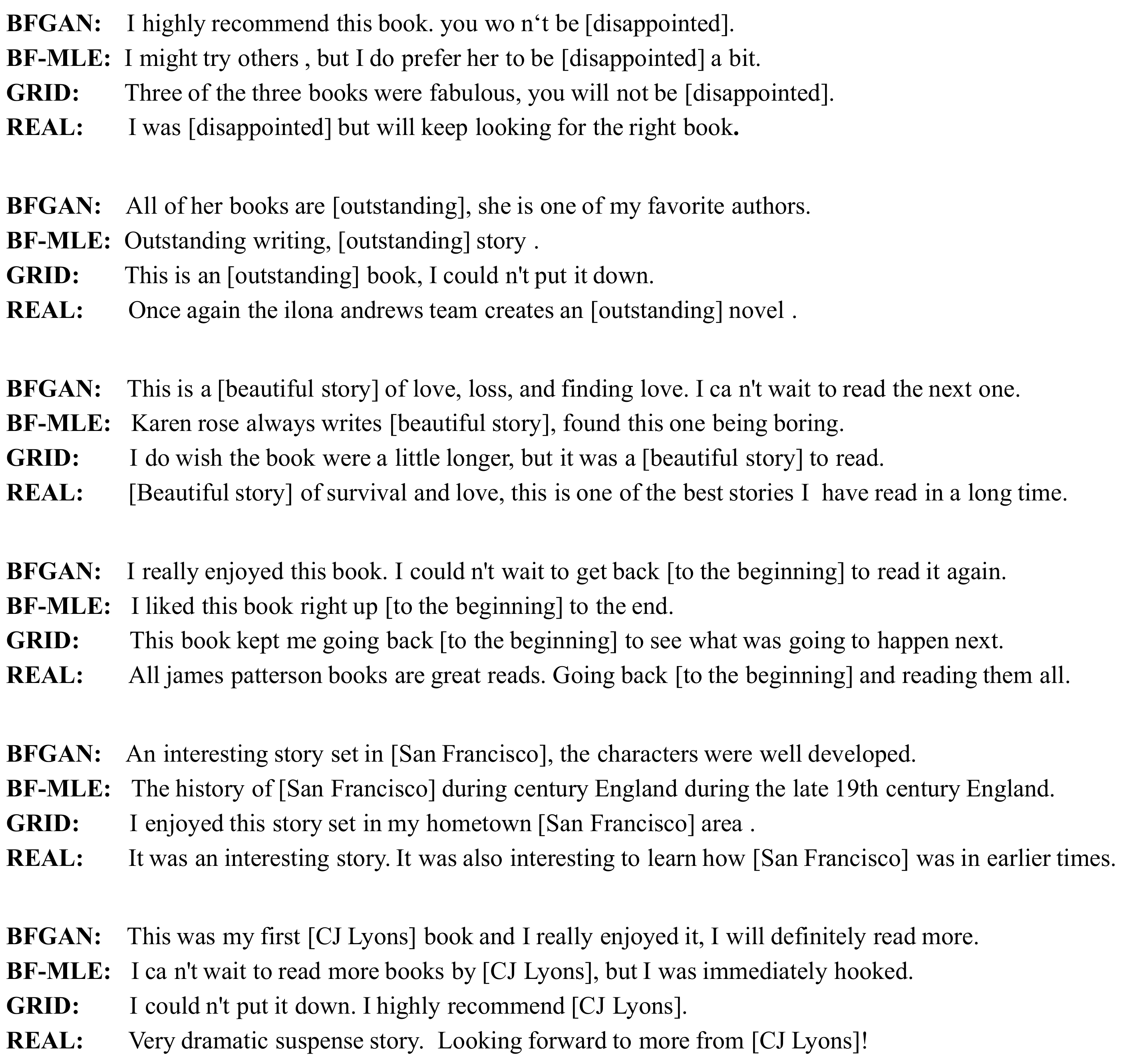}
   \caption{Some generated examples on APRC. GRID: grid beam search, REAL: real samples. We use [$\cdot$] to highlight the lexical constraints.}
   \label{fig:example_aprc}
\end{figure}

\subsection{Samples}
We show some generated examples of different models trained on ARPC in Figure \ref{fig:example_aprc}.

\subsection{Visualization of Attentions}
Figure \ref{fig:visual_aprc} provides four visualization samples of the attention weights in the BFGAN trained on APRC dataset. We highlight in red four output words for four time steps. For each highlighted generated word, we visualize the attention weights of each previous generated word, where darker shades indicate larger attention weights.

\end{document}